\documentclass[11pt,a4paper]{article}
\usepackage[hyperref]{eacl2021}
\usepackage{times}
\usepackage{latexsym}
\usepackage{times}
\usepackage{latexsym}
\usepackage{amsmath,amsthm,amssymb}
\usepackage{graphicx}
\usepackage{enumitem}
\usepackage{setspace}
\usepackage{color}
\usepackage[normalem]{ulem} 
\usepackage{xcolor}
\usepackage{subfigure} 
\theoremstyle{definition}

\usepackage{multirow}
\usepackage{subfigure} 

\usepackage{ textcomp }
\usepackage{ gensymb }
\usepackage{ wasysym }
\usepackage[many]{tcolorbox}
\usepackage{tikz}

\usepackage{booktabs} 
\usepackage{microtype}
\usepackage[table]{colortbl}

\aclfinalcopy 
\def\aclpaperid{375} 

\newcommand{\squishlist}{
	\begin{list}{$\bullet$}
		{ \setlength{\itemsep}{0pt}
			\setlength{\parsep}{3pt}
			\setlength{\topsep}{3pt}
			\setlength{\partopsep}{0pt}
			\setlength{\leftmargin}{1.5em}
			\setlength{\labelwidth}{1em}
			\setlength{\labelsep}{0.5em} } }
	
	\newcommand{\squishlisttwo}{
		\begin{list}{$\bullet$}
			{ \setlength{\itemsep}{0pt}
				\setlength{\parsep}{0pt}
				\setlength{\topsep}{0pt}
				\setlength{\partopsep}{0pt}
				\setlength{\leftmargin}{2em}
				\setlength{\labelwidth}{1.5em}
				\setlength{\labelsep}{0.5em} } }
		
		\newcommand{\squishend}{
	\end{list}  }

\title{ Hierarchical Multi-head Attentive Network \\ for Evidence-aware Fake News Detection}

\author{Nguyen Vo \\
  Worcester Polytechnic Institute \\
  Computer Science Department \\
  Worcester, MA, USA, 01609 \\
  \texttt{nkvo@wpi.edu} \\\And
  Kyumin Lee \\
  Worcester Polytechnic Institute \\
  Computer Science Department \\
  Worcester, MA, USA, 01609 \\
  \texttt{kmlee@wpi.edu} \\}

\begin{document}
\maketitle

\begin{abstract}
	The widespread of fake news and misinformation in various domains ranging from politics, economics to public health has posed an urgent need to automatically fact-check information. A recent trend in fake news detection is to utilize evidence from external sources. However, existing evidence-aware fake news detection methods focused on either only word-level attention or evidence-level attention, which may result in suboptimal performance. In this paper, we propose a Hierarchical Multi-head Attentive Network to fact-check textual claims. Our model jointly combines multi-head word-level attention and multi-head document-level attention, which aid explanation in both word-level and evidence-level. Experiments on two real-word datasets show that our model outperforms seven state-of-the-art baselines. Improvements over baselines are from 6\% to 18\%. Our source code and datasets are released at \texttt{\url{https://github.com/nguyenvo09/EACL2021}}.
	
\end{abstract}

\section{Introduction}
\label{sec:introduction}
The proliferation of biased news, misleading claims, disinformation and fake news has caused heightened negative effects on modern society in various domains ranging from politics, economics to public health. A recent study showed that maliciously fabricated and partisan stories possibly caused citizens' misperception about political candidates \cite{allcott2017social} during the 2016 U.S. presidential elections. In economics, the spread of fake news has manipulated stock price \cite{kogan2019fake}. For example, \$139 billion was wiped out when the Associated Press (AP)'s hacked Twitter account posted rumor about White House explosion with Barack Obama's injury. Recently, misinformation has caused infodemics in public health \cite{Ashoka2020} and even led to people's fatalities in the physical world \cite{WhatsAppIndia}.


To reduce the spread of misinformation and its detrimental influences, many fact-checking systems have been developed to fact-check textual claims. It is estimated that the number of fact-checking outlets has increased 400\% in 60 countries since 2014 \cite{poynter}. Several fact-checking systems such as \textit{snopes.com} and \textit{politifact.com} are widely used by both online users and major corporations. Facebook \cite{FacebookMark95} recently incorporated third-party fact-checking sites to social media posts and Google integrated fact-checking articles to their search engine \cite{wang2018relevant}. These fact-checking systems debunk claims by manually assess their credibility based on collected webpages used as evidence. However, this manual process is laborious and unscalable to handle the large volume of produced false claims on communication platforms. Therefore, in this paper, our goal is to build an automatic fake news detection system to fact-check textual claims based on collected evidence to speed up fact-checking process of the above fact-checking sites.

To detect fake news, researchers proposed to use linguistics and textual content \cite{castillo2011information,zhao2015enquiring,liu2015real}. Since textual claims are usually deliberately written to deceive readers, it is hard to detect fake news by solely relying on the content claims. Therefore, multiple works utilized other signals such as temporal spreading patterns \cite{liu2018early}, network structures \cite{wu2018tracing,vo2018rise,shu2020hierarchical} and users' feedbacks \cite{vo2019learning,shu2019defend,vo2020standing}. However, limited work used external webpages as documents which could provide interpretive explanation to users. Several recent work \cite{popat2018declare,ma2019sentence,vo2020facts} started to utilize documents to fact-check textual claims. \citet{popat2018declare} used word-level attention in documents but treated all documents with equal importance whereas \citet{ma2019sentence} only focused on which documents are more crucial without considering what words help explain credibility of textual claims.

Observing drawbacks of the existing work, we propose Hierarchical Multi-head Attentive Network which jointly utilizes word attention and evidence attention. Overall semantics of a document may be generated by multiple parts of the document. Therefore, we propose a multi-head word attention mechanism to capture different semantic contributions of words to the meaning of the documents.
Since a document may have different semantic aspects corresponding to various information related to credibility of a claim, we propose a multi-head document-level attention mechanism to capture contributions of the different semantic aspects of the documents. In our attention mechanism, we also use speakers and publishers information to further improve effectiveness of our model.
To our knowledge, our work is the first applying multi-head attention mechanism for both words and documents in evidence-aware fake news detection. Our work makes the following contributions:



\squishlist
\item We propose a novel hierarchical multi-head attention network which jointly combines word attention and evidence attention for evidence-aware fake news detection.  
\item We propose a novel multi-head attention mechanism to capture important words and evidence.
\item Experiments on two public datasets demonstrate the effectiveness and generality of our model over state-of-the-art fake news detection techniques. 
\squishend

\section{Related Work}

Many methods have been proposed to detect fake news in recent years. These methods can be placed into three groups: (1) human-based fact-checking sites (e.g. Snopes.com, Politifact.com), (2) machine learning based methods and (3) hybrid systems (e.g. content moderation on social media sites). In machine-learning-based methods, researchers mainly used linguistics and textual content \cite{zellers2019defending,zhao2015enquiring,wang2017liar,shu2019defend}, temporal spreading patterns \cite{liu2018early}, network structures \cite{wu2018tracing,vo2018rise,you2019attributed}, users' feedbacks \cite{vo2019learning,shu2019defend} and multimodal signals \cite{gupta2013faking,vo2020facts}. Recently, researchers focus on fact-checking claims based on evidence from different sources. \citet{thorne2017extensible} and \citet{vlachos2015identification} fact-check claims using subject-predicate-object triplets extracted from knowledge graph as evidence. \citet{chen2019tabfact} assess claims' credibility using tabular data. Our work is closely related to fact verification task \cite{thorne2018fever,nie2018combining,soleimani2020bert} which aims to classify a pair of a claim and an evidence extracted from Wikipedia into three classes: \textit{supported}, \textit{refuted}, or \textit{not enough info}. For fact verification task, \citet{nie2018combining} used ELMo \cite{peters2018deep} to extract contextual embeddings of words and used a modified ESIM model \cite{chen2017enhanced}. \citet{soleimani2020bert} used BERT model \cite{devlin2018bert} to retrieve and verify claims. \citet{zhou2019gear} used graph based models for semantic reasoning.
Our work is different from these work since our goal is to classify a pair of a claim and a list of relevant evidence into \textit{true} or \textit{false}. 

Our work is close to existing work about evidence-aware fake news detection \cite{popat2018declare,ma2019sentence,wuevidence,mishra2019sadhan}. \citet{popat2018declare} used an average pooling layer to derive claims' representation to attend to words in evidence, \citet{mishra2019sadhan} focused on words and sentences in each evidence, and \citet{ma2019sentence} proposed a semantic entailment model to attend to important evidence. However, to the best of our knowledge, our work is the first jointly using multi-head attention mechanisms to focus on important words in each evidence and important evidence from a set of relevant articles. Our attention mechanism is different from these work since we use multiple attention heads to capture different semantic contributions of words and evidence.
%




\section{Problem Statement}
We denote an evidence-based fact-checking dataset $\mathcal{C}$ as a collection of tuples $(c, s, \mathcal{D}, \mathcal{P})$ where $c$ is a textual claim originated from a speaker $s$, $\mathcal{D} = \{d_i\}_{i=1}^k$ is a collection of $k$ documents\footnote{We use the term ``documents'', ``articles'', and ``evidence'' interchangeably.} relevant to the claim $c$ and $\mathcal{P}=\{p_i\}_{i=1}^k$ is the corresponding publishers of documents in $\mathcal{D}$. Note, $|\mathcal{D}| = |\mathcal{P}|$. Our goal is to classify each tuple $(c, s, \mathcal{D}, \mathcal{P})$ into a pre-defined class (i.e. true news/fake news). 


\begin{figure*}[t]
	\includegraphics[width=1\textwidth, height=2.8in,trim={0.1cm 1.5cm 0.9cm 0.8cm},clip]{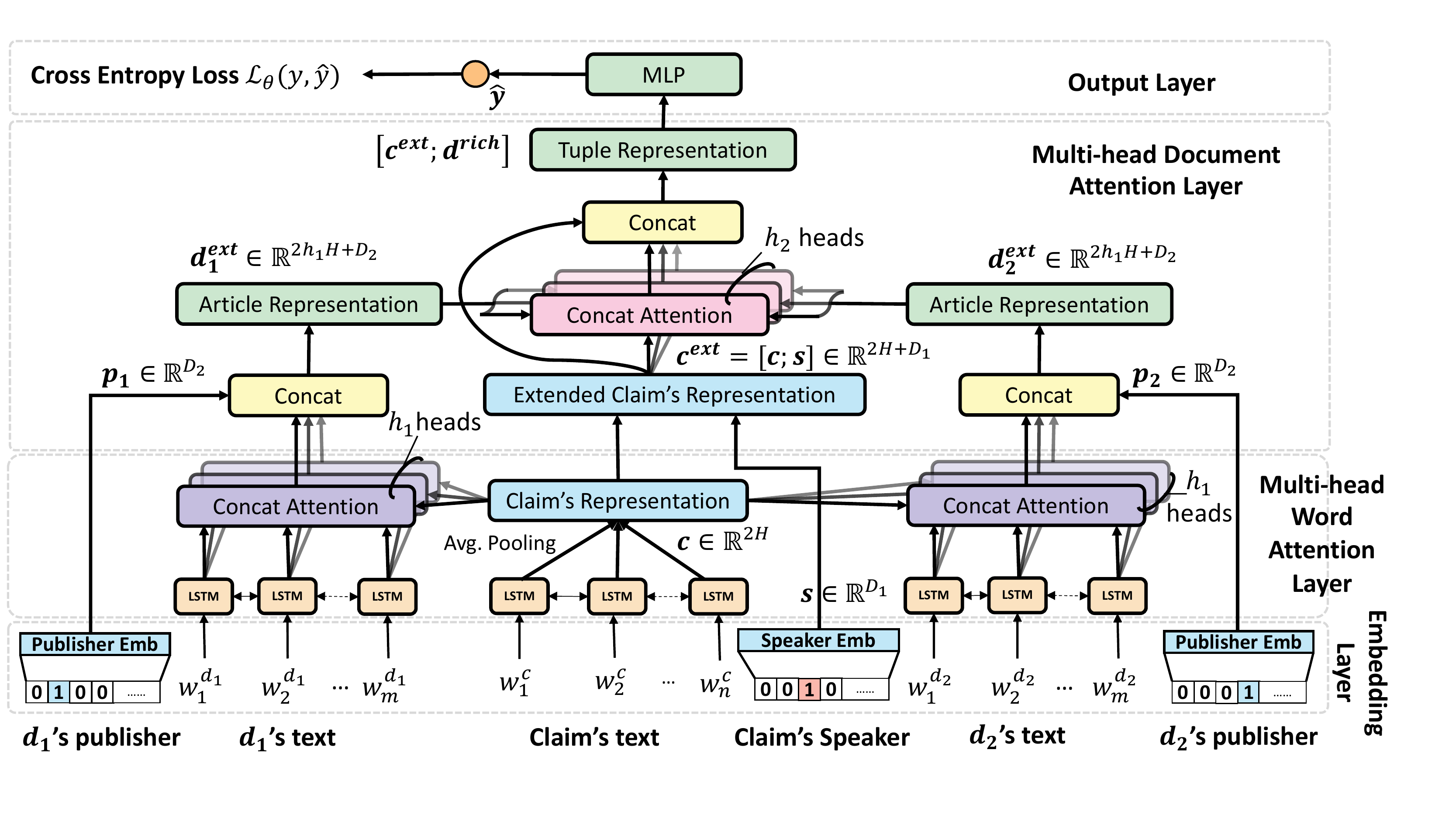}
	\caption{The architecture of our proposed model \textbf{MAC} in which we show a claim $c$, two associated relevant articles $d_1$ and $d_2$ and sources of the claim and the two documents. $h_1$ and $h_2$ are the number of heads of word-level attention and document-level attention respectively.}
	\label{fig:novel_model1}
\end{figure*}
\section{Framework}
In this section, we describe our Hierarchical \underline{\textbf{M}}ulti-head \underline{\textbf{A}}ttentive Network for Fact-\underline{\textbf{C}}hecking (\textbf{MAC}) which jointly considers word-level attention and document-level attention. Our framework consists of four main components: (1) embedding layer, (2) multi-head word attention layer, (3) multi-head document attention layer and (4) output layer. These components are illustrated in Fig. \ref{fig:novel_model1} where we show a claim and two documents as an example.

\subsection{Embedding Layer}
Each claim $c$ is modeled as a sequence of $n$ words $[w_1^c,w_2^c,...,w_n^c]$ and $d_i$ is viewed as another sequence of $m$ words $[w_1^d,w_2^d,...,w_m^d]$. Each word $w_i^c$ and $w_j^d$ will be projected into D-dimensional vectors $\textbf{e}^c_i$ and $\textbf{e}_j^d$ respectively by an embedding matrix $\textbf{W}_e \in \mathbb{R}^{V\times D}$ where $V$ is the vocabulary size. 
Each speaker $s$ and publisher $p_i$ modeled as one-hot vectors are transformed into dense vectors $\textbf{s} \in \mathbb{R}^{D_1}$ and $\textbf{p}_i \in \mathbb{R}^{D_2}$ respectively by using two matrices  $\textbf{W}_{s} \in \mathbb{R}^{S\times D_1}$ and $\textbf{W}_{p} \in \mathbb{R}^{P\times D_2}$, where $S$ and $P$ are the number of speakers and publishers in a training set respectively. Both $\textbf{W}_{s}$ and $\textbf{W}_{p} $ are uniformly initialized in $[-0.2, 0.2]$. Note that, both matrices $\textbf{W}_{s}$ and $\textbf{W}_{p} $ are jointly learned with other parameters of our MAC. 
\subsection{Multi-head Word Attention Layer}
\label{sec:word_attention}
\setlength{\abovedisplayskip}{6pt}
\setlength{\belowdisplayskip}{6pt}
We input word embeddings $\textbf{e}^c_i$ of the claim $c$ into a bidirectional LSTM \cite{graves2005bidirectional} which helps generate contextual representation $\textbf{h}_i$ of each token as follows: $\textbf{h}^c_i = [\overleftarrow{\textbf{h}_i};\overrightarrow{\textbf{h}_i}]\in \mathbb{R}^{2H}$, where  $\overleftarrow{\textbf{h}}_i$ and $\overrightarrow{\textbf{h}}_i$ are hidden states in forward and backward pass of the BiLSTM, symbol $;$ means concatenation and $H$ is hidden size.
We derive claim's representation in $\mathbb{R}^{2H}$ by an average pooling layer as follows:
\begin{equation}
\textbf{c} = {1 \over n} \sum_{i=1}^n \textbf{h}^c_i
\label{eq:claim_represation_text_only}
\end{equation}

Applying a similar process on the top of each document $d_i$ with a different BiLSTM, we have contextual representation $\textbf{h}_j^d\in \mathbb{R}^{2H}$ for each word in $d_i$. After going through BiLSTM, $d_i$ is modeled as matrix $\textbf{H} = [\textbf{h}_1^d \oplus \textbf{h}_2^d\oplus ...\oplus\textbf{h}_m^d] \in \mathbb{R}^{m\times 2H}$ where $\oplus$ denotes stacking. 

To understand what information in a document helps us fact-check a claim, we need to guide our model to focus on crucial keywords or phrases of the document. Drawing inspiration from \cite{luong2015effective}, we firstly replicate vector $\textbf{c}$ (Eq.\ref{eq:claim_represation_text_only}) $m$ times to create matrix $\textbf{C}_1\in \mathbb{R}^{m\times 2H}$ and propose an attention mechanism to attend to important words in the document $d_i$ as follows:
\begin{equation}
\textbf{a}_1 =  softmax\big( \tanh\big([\textbf{H}; \textbf{C}_1] \cdot \textbf{W}_1\big) \cdot \textbf{w}_{2} \big)
\label{eq:word_level_att_single_head}
\end{equation}
where  $\textbf{w}_{2} \in \mathbb{R}^{a_1}$, $\textbf{W}_1 \in \mathbb{R}^{4H \times a_1}$, $[\textbf{H};\textbf{C}_1]$ is concatenation of two matrices on the last dimension and $\textbf{a}_1 \in \mathbb{R}^m$ is attention distribution on $m$ words.
However, the overall semantics of the document might be generated by multiple parts of the document \cite{lin2017structured}. Therefore, we propose a multi-head word attention mechanism to capture different semantic contributions of words by extending vector $\textbf{w}_{2}$ into a matrix $\textbf{W}_{2} \in \mathbb{R}^{a_1 \times h_1}$ where $h_1$ is the number of attention heads shown in Fig.~\ref{fig:novel_model1}. We modify Eq.~\ref{eq:word_level_att_single_head} as follows:
\begin{equation}
\textbf{A}_1 =  softmax_{col}\big( \tanh([\textbf{H}; \textbf{C}_1] \cdot \textbf{W}_1) \cdot \textbf{W}_{2} \big)
\label{eq:word_level_att_multi_head}
\end{equation}
where $\textbf{A}_1 \in \mathbb{R}^{m\times h_1}$ and each column of $\textbf{A}_1$ has been normalized by the softmax operation. Intuitively, $\textbf{A}_1$ stands for $h_1$ different attention distributions on top of $m$ words of the document $d_i$, helping us capture different aspects of the document. After computing $\textbf{A}_1$, we derive representation of document $d_i$ as follows:
\begin{equation}
\textbf{d}_i = flatten(\textbf{A}^T_1\cdot \textbf{H})
\label{eq:single_doc_representation}
\end{equation}
where $\textbf{d}_i  \in \mathbb{R}^{h_1 2H}$ and function \textit{flatten(.)} flattens $\textbf{A}^T_1 \cdot \textbf{H}$ into a vector. We also implemented a more sophisticated multi-head attention in \cite{vaswani2017attention} but did not achieve good results.

\subsection{Multi-head Document Attention Layer}
This layer consists of three components as follows: (1) extending representations of claims, (2) extending representations of evidence and (3) multi-head document attention mechanism.

\noindent\textbf{Extending representations of claims.}
So far the representation of the claim $\textbf{c}$ (Eq.~\ref{eq:claim_represation_text_only}) is only from textual content. In reality, a speaker who made a claim may impact credibility of the claim. For example, claims from some politicians are controversial and inaccurate \cite{allcott2017social}. Therefore, we enrich vector $\textbf{c}$ by concatenating it with speaker's embedding $\textbf{s}$ to generate $\textbf{c}^{ext} \in \mathbb{R}^{x}$, where $x=2H+D_1$ as shown in Eq.~\ref{eq:claim_rich}.
\begin{equation}
\textbf{c}^{ext} = [\textbf{c};\textbf{s}] \in \mathbb{R}^{x}
\label{eq:claim_rich}
\end{equation}

\noindent\textbf{Extending representations of evidence.}
Intuitively, an article published by \textit{nytimes.com} might be more reliable than a piece of news published by \textit{breitbart.com} which is known to be a less credible site. Therefore, to capture more information, we further enrich representations of evidence with publishers' information by concatenating $\textbf{d}_i$ (Eq.~\ref{eq:single_doc_representation}) with its publisher's embedding $\textbf{p}_i$ as follows:
\begin{equation}
	\textbf{d}^{ext}_i = [\textbf{d}_i;\textbf{p}_i] \in \mathbb{R}^{y}
	\label{eq:doc_rich}
\end{equation}
where $y=2h_1H + D_2$. 
From Eq.~\ref{eq:doc_rich}, we can generate representations of $k$ relevant articles and stack them as shown in Eq.~\ref{eq:all_doc_representations}.
\begin{equation}
\textbf{D} = [\textbf{d}^{ext}_1 \oplus...\oplus\textbf{d}^{ext}_k] \in \mathbb{R}^{k\times y}
\label{eq:all_doc_representations}
\end{equation}


\noindent\textbf{Multi-head Document Attention Mechanism.}
 In real life, a journalist from \textit{snopes.com} and \textit{politifact.com} may use all $k$ articles relevant to the claim $c$ to fact-check it but she may focus on some key articles to determine the verdict of the claim $c$ while other articles may have negligible information. To capture such intuition, we need to downgrade uninformative documents and concentrate on more meaningful articles. Similar to Section \ref{sec:word_attention}, we use multi-head attention mechanism which produces different attention distributions representing diverse contributions of articles toward determining veracity of the claim $c$.

We firstly create matrix $\textbf{C}_2 \in \mathbb{R}^{k\times x}$ by replicating vector $\textbf{c}^{ext}$ (Eq.~\ref{eq:claim_rich}) $k$ times. Secondly, the matrix $\textbf{C}_2$ is concatenated with matrix $\textbf{D}$  (Eq.~\ref{eq:all_doc_representations}) on the last dimension of the two matrices denoted as $[\textbf{D}; \textbf{C}_2] \in \mathbb{R} ^ {k\times (x + y)}$.

Our proposed multi-head document-level attention mechanism applies $h_2$ different attention heads as shown in Eq.~\ref{eq:document_level_att_multi_head}.
\begin{equation}
\textbf{A}_2 =  softmax_{col}( \tanh([\textbf{D}; \textbf{C}_2] \cdot \textbf{W}_3) \cdot \textbf{W}_{4} )
\label{eq:document_level_att_multi_head}
\end{equation}
where $\textbf{W}_3 \in \mathbb{R}^{(x+y)\times a_2}$, $\textbf{W}_{4}\in \mathbb{R}^{a_2\times h_2}$. 
The matrix $\textbf{A}_2 \in \mathbb{R}^{k\times h_2}$, where each of its column is normalized by the softmax operator, is a collection of $h_2$ different attention distributions on $k$ documents. Using attention weights, we can generate attended representation of $k$ evidence denoted as $\textbf{d}^{rich}\in \mathbb{R}^{h_2y}$ as shown in Eq.~\ref{eq:d_rich}.
\begin{equation}
\textbf{d}^{rich} = flatten(\textbf{A}^T_2\cdot \textbf{D})
\label{eq:d_rich}
\end{equation}
where \textit{flatten(.)} function flattens $\textbf{A}^T_2\cdot \textbf{D}$ into a vector. We finally generate representation of a tuple $(c,s,\mathcal{D},\mathcal{P})$ by concatenating vector $\textbf{c}^{ext}$ (Eq.~\ref{eq:claim_rich}) and vector $\textbf{d}^{rich}$ (Eq.~\ref{eq:d_rich}), denoted as $[\textbf{c}^{ext};\textbf{d}^{rich}]$.

To the best of our knowledge, our work is the first work utilizing multi-head attention mechanism integrated with speakers and publishers information to capture various semantic contributions of evidence toward fact-checking process.
%
%



\subsection{Output Layer}
In this layer, we input tuple representation $[\textbf{c}^{ext};\textbf{d}^{rich}]$ into a multilayer perceptron (MLP) to compute probability $\hat{y}$ that the claim $c$ is a true news as follows:
\begin{equation}
\hat{y} = \sigma\big(\textbf{W}_6 \cdot \big(\textbf{W}_5\cdot [\textbf{c}^{ext};\textbf{d}^{rich}] + \textbf{b}_5 \big) + \textbf{b}_6\big)
\end{equation}
where $\textbf{W}_5, \textbf{W}_6, \textbf{b}_5, \textbf{b}_6$ are weights and biases of the MLP, and $\sigma(.)$ is the sigmoid function. We optimize our model by minimizing the standard cross-entropy as shown on the top of Fig.~\ref{fig:novel_model1}.
\begin{equation}
\mathcal{L}_{\theta}(y, \hat{y}) = -\big(y\log{\hat{y}} + (1 - y)\log({1-\hat{y}})\big)
\end{equation}
where $y\in \{0,1\}$ is the ground truth label of a tuple $(c, s, \mathcal{D}, \mathcal{P})$. During training, we sample a mini batch of 32 tuples and compute average loss from the tuples. 

\section{Experiments}
\subsection{Datasets}
We employed two public datasets released by \cite{popat2018declare}. Each of these datasets is a collections of tuples $(c, s, \mathcal{D}, \mathcal{P}, y)$ where each textual claim $c$ and its credible label $y$ are collected from two major fact-checking websites \texttt{snopes.com} and \texttt{politifact.com}. The articles pertinent to the claim $c$ are retrieved by using search engines. Each Snopes claim was labeled as \textit{true} or \textit{false} while in Politifact, there were originally six labels: \textit{true}, \textit{mostly true}, \textit{half true}, \textit{false}, \textit{mostly false}, \textit{pants on fire}. Following \cite{popat2018declare}, we merge \textit{true}, \textit{mostly true} and \textit{half true} into \textit{true claims} and the rest are into \textit{false claims}. Details of our datasets are presented in Table \ref{tbl:dataset_statistics}. Note that Snopes does not have speakers' information.


\subsection{Baselines}
\begin{table}[t]
	\centering
	\caption{Statistics of our experimental datasets}
	\begin{tabular}{lcc}
		\toprule[1pt]
		& Snopes & PolitiFact \\ \hline
		True claims  & 1,164  & 1,867      \\ \hline
		False claims & 3,177  & 1,701      \\ \hline
		$|$Speakers$|$   & N/A     & 664        \\ \hline
		$|$Documents$|$  & 29,242 & 29,556     \\ \hline
		$|$Publishers$|$ & 12,236 & 4,542      \\ \bottomrule[1pt]
	\end{tabular}
	\vspace{-10pt}
	\label{tbl:dataset_statistics}
\end{table}
We compare our MAC model with seven state-of-the-art baselines divided into two groups. The first group of the baselines only used textual content of claims, and the second group of the baselines utilized relevant articles to fact-check textual claims. A related method \cite{mishra2019sadhan} used subject information of articles (e.g. politics, entertainment), which was not available in our datasets. We tried to compare with it but achieved poor results perhaps due to missing information. Therefore, we do not report its result in this paper. Details of the baselines are shown as follows:

\noindent\textbf{Using only claims' text: }
\squishlist
\item \textbf{BERT} \cite{devlin2018bert} is a pre-trained language model achieving state-of-the-art results on many NLP tasks. The representation of [CLS] token is inputted to a trainable linear layer to classify claims.
\item \textbf{LSTM-Last} is a model proposed in \cite{rashkin2017truth}. \textbf{LSTM-Last} takes the last hidden state of the LSTM as representations of claims. These representations will be inputted to a linear layer for classification.
\item \textbf{LSTM-Avg} is another model proposed in \cite{rashkin2017truth} which used an average pooling layer on top of hidden states to derive representations of claims.
\item \textbf{CNN} \cite{wang2017liar} is a state-of-the-art model which applied 1D-convolutional neural network on word vectors of claims.
\squishend


\noindent\textbf{Using both claims' text and articles' text: }
\squishlist
\item \textbf{DeClare} \cite{popat2018declare} computes credibility score of each pair of a claim $c$ and a document $d_i$. The overall credible rating is averaged from all $k$ relevant articles.
\item \textbf{HAN} \cite{ma2019sentence} is a hierarchical attention network based on representations of relevant documents. It uses attention mechanisms to determine which document is more important without considering which word in a document should be focused on.
\item \textbf{NSMN} \cite{nie2018combining} is a state-of-the-art model designed to determine stance of a document $d_i$ with respect to claim $c$. We apply NSMN on our dataset by predicting score of each pair $(c, d_i)$ and computing average score based on documents in $\mathcal{D}$ same as DeClare.
\squishend

Note that, we also applied BERT, LSTM-Last, LSTM-Avg and CNN by using both claims' text and articles' text. For each of these baselines, we concatenated a claim's text and a document's text, and input the concatenated content into the baseline to compute likelihood that the claim is fake news. We computed average probability based on all documents of the claim and used it as final prediction. However, we did not observe considerable improvements of these baselines. In addition to deep-learning-based baselines, we compared our MAC with other feature-based techniques (e.g. SVM). As expected, these traditional techniques had inferior performance compared with neural models. Therefore, we only report the seven baselines' performance. 

\begin{table*}[t]
	\caption{Performance of MAC and baselines on Snopes dataset. MAC outperforms baselines significantly with p-value$<$0.05 by one-sided paired Wilcoxon test.}
	\label{tbl:performance_on_Snopes_optimizing_f1_macro_5_folds}
	\resizebox{1.0\linewidth}{!} {	
		
		\begin{tabular}{lllll|lll|lll}
			\toprule[1pt]
			\multicolumn{1}{c}{\multirow{2}{*}{\begin{tabular}[c]{@{}c@{}}Method\\ Types\end{tabular}}}                                  & \multicolumn{1}{c}{\multirow{2}{*}{Methods}} &                            &                            &                             & \multicolumn{3}{c|}{True News as Positive}                                               & \multicolumn{3}{c}{Fake News as Positive}                                           \\ \cline{3-11}
			\multicolumn{1}{c}{}                                                                                                         & \multicolumn{1}{c}{}                         & \multicolumn{1}{c}{AUC}    & \multicolumn{1}{c}{F1 Macro} & \multicolumn{1}{c|}{F1 Micro} & \multicolumn{1}{c}{F1}      & \multicolumn{1}{c}{Precision} & \multicolumn{1}{c|}{Recall}  & \multicolumn{1}{c}{F1}     & \multicolumn{1}{c}{Precision} & \multicolumn{1}{c}{Recall} \\ \hline
			
			\multicolumn{1}{l|}{\multirow{4}{*}{\begin{tabular}[c]{@{}l@{}}Using only \\ claims' text\end{tabular}}}                     & \multicolumn{1}{l|}{BERT}                    & 0.60852                    & 0.56096                    & 0.69806                     & 0.31574                     & 0.40318                     & 0.26050                      & 0.80618                    & 0.76011                    & 0.85839                    \\
			\multicolumn{1}{l|}{}                                                                                                        & \multicolumn{1}{l|}{LSTM-Avg}                & 0.69124                    & 0.62100                    & 0.71877                     & 0.42953                     & 0.48415                     & 0.39692                      & 0.81246                    & 0.79139                    & 0.83671                    \\
			\multicolumn{1}{l|}{}                                                                                                        & \multicolumn{1}{l|}{LSTM-Last}               & 0.70142                    & 0.63122                    & 0.72415                     & 0.44650                     & 0.48935                     & 0.41412                      & 0.81594                    & 0.79594                    & 0.83776                    \\
			\multicolumn{1}{l|}{}                                                                                                        & \multicolumn{1}{l|}{TextCNN}                 & 0.70537                    & 0.63081                    & 0.72005                     & 0.45001                     & 0.48164                     & 0.43035                      & 0.81160                    & 0.79882                    & 0.82622                    \\ \hline
			\multicolumn{1}{l|}{\multirow{3}{*}{\begin{tabular}[c]{@{}l@{}}Using both\\ claims' text \& \\ articles' text\end{tabular}}} & \multicolumn{1}{l|}{HAN}                     & 0.70365                    & 0.62510                    & 0.72800                     & 0.42884                     & 0.49192                     & 0.38161                      & 0.82136                    & 0.79058                    & 0.85490                    \\
			\multicolumn{1}{l|}{}                                                                                                        & \multicolumn{1}{l|}{NSMN}                    & 0.77270                    & 0.68006                    & 0.76127                     & 0.51954                     & 0.57558                     & 0.48182                      & 0.84058                    & 0.82011                    & 0.86364                    \\
			\multicolumn{1}{l|}{}                                                                                                        & \multicolumn{1}{l|}{DeClare}                 & 0.81036                    & 0.72445                    & 0.78813                     & 0.59250                     & 0.61235                     & 0.58096                      & 0.85640                    & 0.85023                    & 0.86399                    \\ \hline
			\multicolumn{1}{c|}{Ours}                                                                                                    & \multicolumn{1}{l|}{MAC}               & \textbf{0.88715}           & \textbf{0.78660}           & \textbf{0.83316}            & \textbf{0.68738}            & \textbf{0.69975}            & \textbf{0.68601}             & \textbf{0.88581}           & \textbf{0.88617}           & \textbf{0.88706}           \\ \hline
			\multicolumn{2}{l|}{Imprv. over the best baseline}                                                                                                                           & \multicolumn{1}{r}{9.47\%} & \multicolumn{1}{r}{8.58\%} & \multicolumn{1}{r|}{5.71\%} & \multicolumn{1}{r}{16.01\%} & \multicolumn{1}{r}{14.27\%} & \multicolumn{1}{r|}{18.08\%} & \multicolumn{1}{r}{3.43\%} & \multicolumn{1}{r}{4.23\%} & \multicolumn{1}{r}{2.67\%} \\ \bottomrule[1pt]
		\end{tabular}
	}
\end{table*}

\subsection{Experimental Settings}
For each dataset, we randomly select 10\% number of claims from each class to form a validation set, which is used for tuning hyper-parameters. We report 5-fold stratified cross validation results on the remaining 90\% of the data. We train our model and baselines on 4-folds and test them on the remaining fold. We use AUC, macro/micro F1, class-specific F1, Precision and Recall as evaluation metrics. To mitigate overfitting and reduce training time, we early stop training process on the validation set when F1 macro on the validation data continuously decreases in 10 epochs. When we get the same F1 macro between consecutive epochs, we rely on AUC for early stopping.

For fair comparisons, we use Adam optimizer \cite{kingma2014adam} with learning rate 0.001 and regularize parameters of all methods with $\ell_2$ norm and weight decay $\lambda = 0.001$. As the maximum lengths of claims and articles in words are 30 and 100 respectively for both datasets, we set $n=30$ and $m=100$. For HAN and our model, we set $k=30$ since the number of articles for each claim is at most 30 in both datasets. Batch size is set to 32 and we trained all models until convergence. We tune all models including ours with hidden size $H$ chosen from $\{64, 128, 300\}$, pre-trained word-embeddings are from Glove \cite{pennington2014glove} with $D=300$. Both $D_1$ and $D_2$ are tuned from $\{128, 256\}$. The number of attention heads $h_1$ and $h_2$ is chosen from $\{1,2,3,4,5\}$, $a_1$ and $a_2$ are equal to $2\times H$. In addition to Glove, we also utilized contextual embeddings from pre-trained language models such as ELMo and BERT but achieved comparable performances. We implemented all methods in PyTorch 0.4.1 and run experiments on an NVIDIA GTX 1080. 

\begin{table*}[t]
	\caption{Performance of MAC and baselines on PolitiFact dataset. MAC outperforms baselines with statistical significance level p-value$<$0.05 by one-sided paired Wilcoxon test.}
	\label{tbl:performance_on_Politifact_optimize_based_on_F1_macro_5_folds}
	\resizebox{1.0\linewidth}{!} {	
		
		\begin{tabular}{lllll|lll|lll}
			\toprule[1pt]
			\multicolumn{1}{c}{\multirow{2}{*}{\begin{tabular}[c]{@{}c@{}}Method\\ Types\end{tabular}}}                                  & \multicolumn{1}{c}{\multirow{2}{*}{Methods}} &                            &                              &                               & \multicolumn{3}{c|}{True News as Positive}                                                & \multicolumn{3}{c}{Fake News as Positive}                                              \\ \cline{3-11}
			\multicolumn{1}{c}{}                                                                                                         & \multicolumn{1}{c}{}                         & \multicolumn{1}{c}{AUC}    & \multicolumn{1}{c}{F1 Macro} & \multicolumn{1}{c|}{F1 Micro} & \multicolumn{1}{c}{F1}     & \multicolumn{1}{c}{Precision} & \multicolumn{1}{c|}{Recall}  & \multicolumn{1}{c}{F1}     & \multicolumn{1}{c}{Precision} & \multicolumn{1}{c}{Recall} \\ \hline
			\multicolumn{1}{l|}{\multirow{4}{*}{\begin{tabular}[c]{@{}l@{}}Using only \\ claims' text\end{tabular}}}                     & \multicolumn{1}{l|}{BERT}                    & 0.58822                    & 0.56021                      & 0.56446                       & 0.56364                    & 0.59206                       & 0.54968                      & 0.55678                    & 0.54354                       & 0.58069                    \\
			\multicolumn{1}{l|}{}                                                                                                        & \multicolumn{1}{l|}{LSTM-Avg}                & 0.65465                    & 0.60564                      & 0.60866                       & 0.61821                    & 0.63192                       & 0.61267                      & 0.59307                    & 0.59046                       & 0.60425                    \\
			\multicolumn{1}{l|}{}                                                                                                        & \multicolumn{1}{l|}{LSTM-Last}               & 0.64289                    & 0.60196                      & 0.60493                       & 0.61703                    & 0.62634                       & 0.61456                      & 0.58690                    & 0.58763                       & 0.59434                    \\
			\multicolumn{1}{l|}{}                                                                                                        & \multicolumn{1}{l|}{TextCNN}                 & 0.65152                    & 0.60380                      & 0.60740                       & 0.61521                    & 0.63010                       & 0.61030                      & 0.59238                    & 0.59049                       & 0.60421                    \\ \hline
			\multicolumn{1}{l|}{\multirow{3}{*}{\begin{tabular}[c]{@{}l@{}}Using both\\ claims' text \& \\ articles' text\end{tabular}}} & \multicolumn{1}{l|}{HAN}                     & 0.63201                    & 0.58655                      & 0.59121                       & 0.59193                    & 0.61502                       & 0.58290                      & 0.58117                    & 0.57573                       & 0.60034                    \\
			\multicolumn{1}{l|}{}                                                                                                        & \multicolumn{1}{l|}{NSMN}                    & 0.64237                    & 0.60211                      & 0.60431                       & 0.61123                    & 0.63051                       & 0.59912                      & 0.59299                    & 0.58213                       & 0.60999                    \\
			\multicolumn{1}{l|}{}                                                                                                        & \multicolumn{1}{l|}{DeClare}                 & 0.70642                    & 0.65213                      & 0.65350                       & 0.67230                    & 0.66548                       & 0.67997                      & 0.63195                    & 0.64053                       & 0.62444                    \\ \hline
			\multicolumn{1}{c|}{Ours}                                                                                                    & \multicolumn{1}{l|}{MAC}                     & \textbf{0.75756}           & \textbf{0.68642}             & \textbf{0.69116}              & \textbf{0.71786}           & \textbf{0.68856}              & \textbf{0.75493}             & \textbf{0.65498}           & \textbf{0.70546}              & \textbf{0.62576}           \\ \hline
			\multicolumn{2}{l|}{Imprv. over the best baseline}                                                                                                                           & \multicolumn{1}{r}{7.24\%} & \multicolumn{1}{r}{5.26\%}   & \multicolumn{1}{r|}{5.76\%}   & \multicolumn{1}{r}{6.78\%} & \multicolumn{1}{r}{3.47\%}    & \multicolumn{1}{r|}{11.02\%} & \multicolumn{1}{r}{3.64\%} & \multicolumn{1}{r}{10.14\%}   & \multicolumn{1}{r}{0.21\%} \\ \bottomrule[1pt]
		\end{tabular}
	}
\end{table*}

\subsection{Performance of MAC and baselines}
We show experimental results of our model and baselines in Tables \ref{tbl:performance_on_Snopes_optimizing_f1_macro_5_folds} and \ref{tbl:performance_on_Politifact_optimize_based_on_F1_macro_5_folds}.
In Table \ref{tbl:performance_on_Snopes_optimizing_f1_macro_5_folds}, MAC outperforms all baselines with significance level $p<0.05$ by using one-sided paired Wilcoxon test on Snopes dataset. MAC achieves the best result when $h_1=5, h_2=2,H=300$ and $D_1=D_2=128$. In Table \ref{tbl:performance_on_Politifact_optimize_based_on_F1_macro_5_folds}, MAC also significantly outperforms all baselines with $p<0.05$ according to one-sided paired Wilcoxon test on PolitiFact dataset. The hyperparameters we selected for MAC are  $h_1=3, h_2=1,H=300$ and $D_1=D_2=128$.

For baselines, BERT is used as a static encoder. We tried to fine tune it but even achieve worse results. This might be because we do not have sufficient data to tune it. For both HAN and DeClare, since both papers do not release their source code, we tried our best to reproduce results from these two models. HAN model derived representation of each document by using the last hidden state of a GRU \cite{chung2014empirical} without any attention mechanism on words to downgrade unimportant words (e.g. stop words), leading to poor representations of documents. Therefore, document-level attention mechanism in HAN model did not perform well. Similar patterns can be observed in two baselines LSTM-Avg and LSTM-Last. DeClare performed best among baselines, indicating the importance of applying word-level attention on words to reduce impact of less informative words.

We can see that our MAC outperforms all baselines in all metrics. When viewing \textit{true news} as \textit{positive class}, our MAC has an average increase of 16.0\% and 7.1\% over the best baselines on Snopes and PolitiFact respectively. We also have an increase of 4.7\% improvements over baselines with a maximum improvements of 10.1\% in PolitiFact when considering \textit{fake news} as \textit{negative class}. In terms of AUC, average improvements of MAC over the baselines are 7.9\% and 6.1\% on Snopes and PolitiFact respectively. Improvements of MAC over baselines can be explained by our multi-head attention mechanism shown in Eq.~\ref{eq:word_level_att_multi_head} and Eq.~\ref{eq:document_level_att_multi_head}. After attending to words in documents, we can generate better representations of documents/evidence, leading to more effective document-level attention compared with HAN model.
\begin{table}[t]
	\caption{Impact of word attention and evidence attention on our MAC in two datasets}
	\resizebox{1.0\columnwidth}{!} {	
		\begin{tabular}{l|cc|cc}
			\toprule[1pt]
			\multicolumn{1}{l|}{\multirow{2}{*}{Methods}} & \multicolumn{2}{c|}{Snopes}                             & \multicolumn{2}{c}{PolitiFact}                        \\ \cline{2-5}
			\multicolumn{1}{c|}{}                         & \multicolumn{1}{l}{AUC} & \multicolumn{1}{l|}{F1 Macro} & \multicolumn{1}{l}{AUC} & \multicolumn{1}{l}{F1 Macro} \\ \hline
			Only Word Att                                 & 0.87278                 & 0.77831                       & 0.74483                 & 0.67818                      \\
			Only Evidence Att                             & 0.82531                 & 0.72885                       & 0.71790                 & 0.65187                      \\
			Word \& Doc Att                               & \textbf{0.88715}                 & \textbf{0.78660 }                      & \textbf{0.75756}                 & \textbf{0.68642}                      \\
			\bottomrule[1pt]
		\end{tabular}
	}
	\label{tbl:ablation_studies_on_word_evd_att}
\end{table}
\begin{table}[t]
	\caption{Impact of speakers and publishers on performance of MAC in two datasets}
		\resizebox{1.0\columnwidth}{!} {	
	\centering
	\begin{tabular}{l|cc|cc}
		\toprule[1pt]
		\multirow{2}{*}{Methods} & \multicolumn{2}{c|}{Snopes}         & \multicolumn{2}{c}{PolitiFact}     \\ \cline{2-5}
		& AUC              & F1 Macro         & AUC              & F1 Macro         \\ \hline
		Text Only                & 0.88186          & 0.77146          & 0.72401          & 0.66844          \\
		Text + Publishers        & \textbf{0.88715} & \textbf{0.78660} & 0.72645          & 0.66984          \\
		Text + Speakers          &     \cellcolor{gray!10}              &      \cellcolor{gray!10}             & 0.75202          & 0.68483          \\
		Text + Pubs + Spkrs      &    \cellcolor{gray!10}               &   \cellcolor{gray!10}                & \textbf{0.75756} & \textbf{0.68642} \\ \bottomrule[1pt]
	\end{tabular}
		}
	\label{tbl:ablation_studies}
\end{table}
\begin{figure}[t]
	
	\subfigure[Snopes]{
		\includegraphics[width=0.47\linewidth, height=1.4in,trim=0 140 30 150,clip]{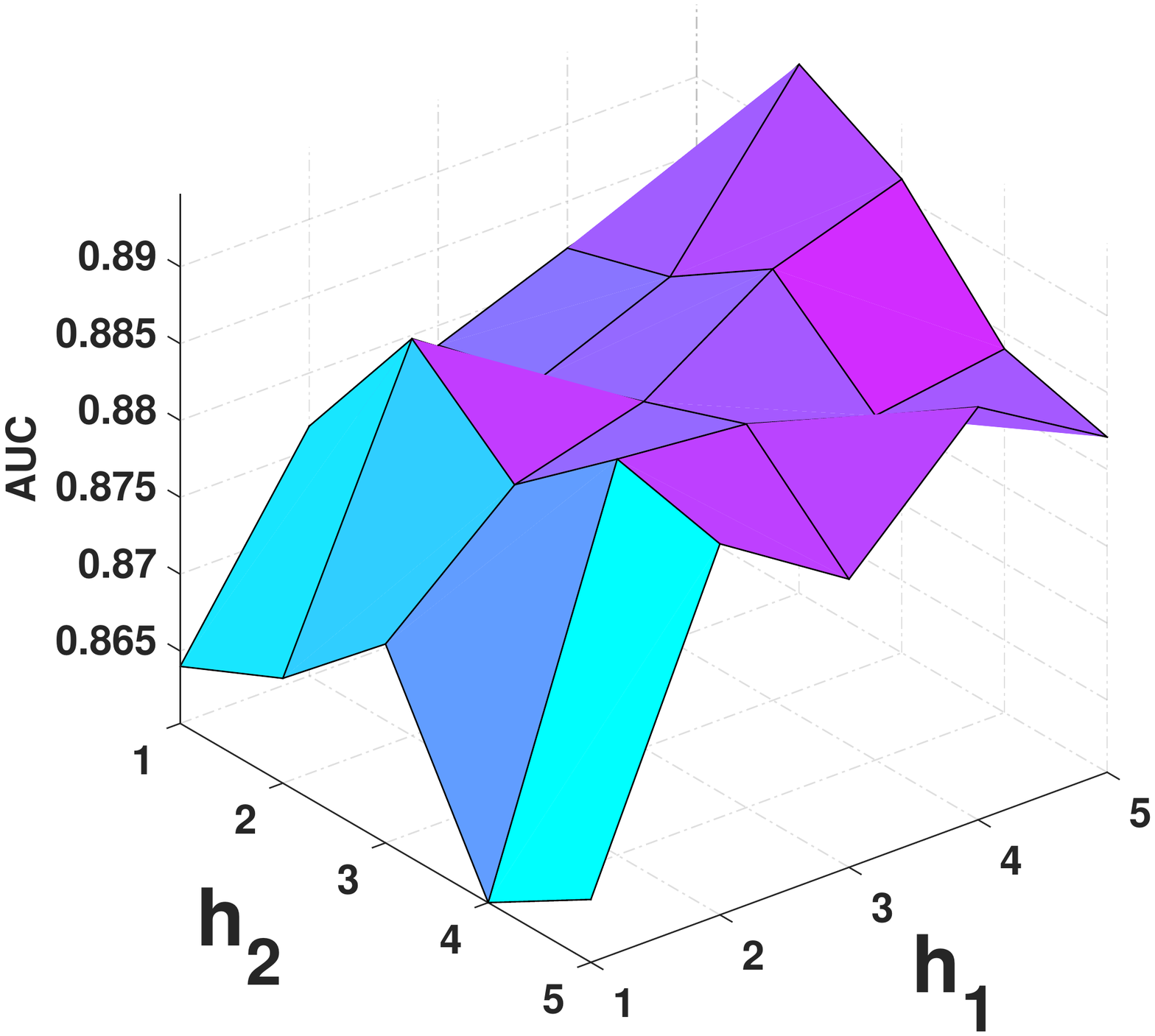}
		\label{fig:snopes_auc_h1_h2}
	}
	\subfigure[PolitiFact ]{
		\includegraphics[width=0.47\linewidth, height=1.4in,trim=55 160 80 200,clip]{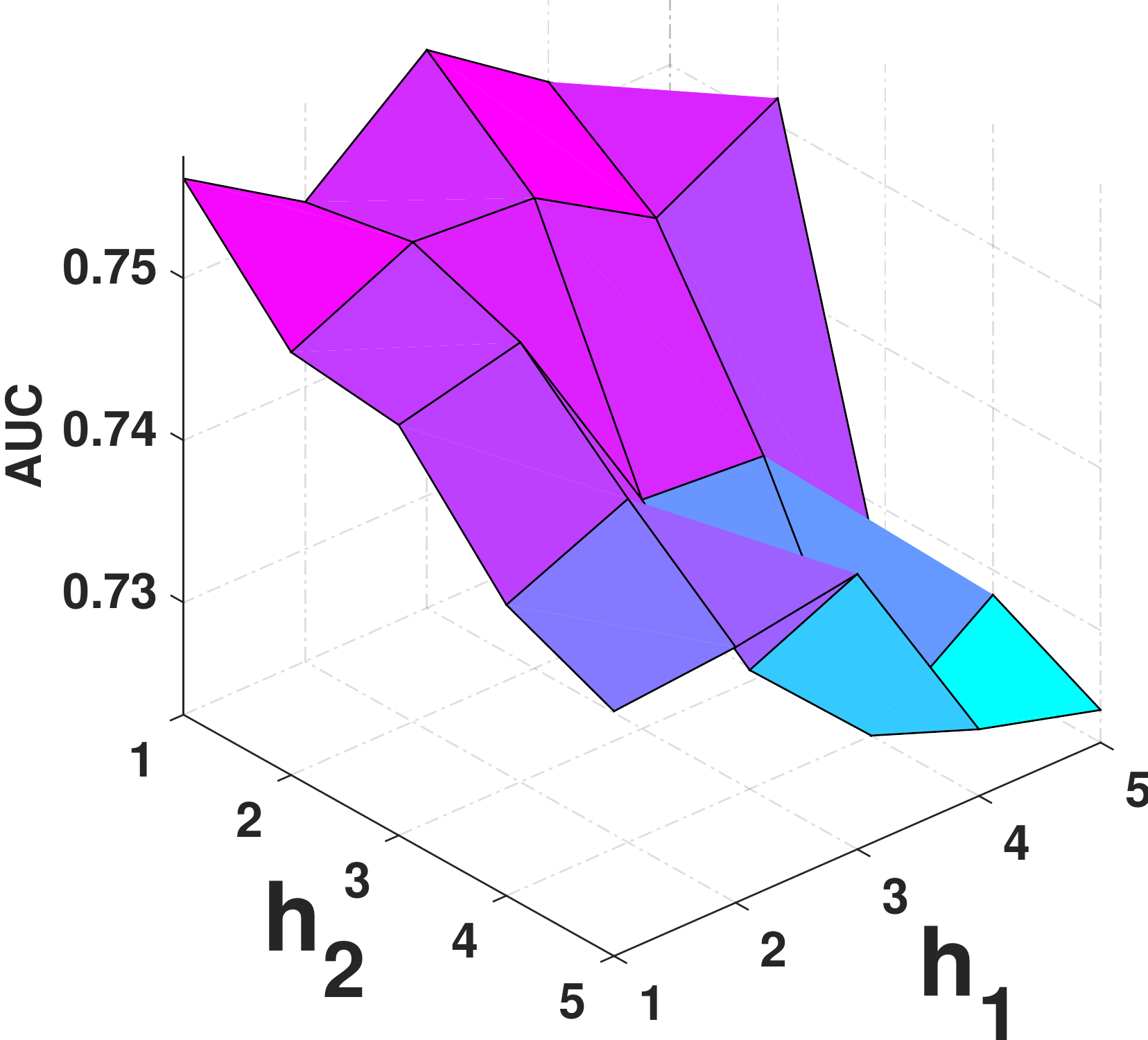}
		\label{fig:politifact_auc_h1_h2}
	}
	\caption{Sensitivity of MAC with respect to number of heads in word-level attention $h_1$ and the number of heads in document-level attention $h_2$}
	\vspace{-10pt}
\end{figure}

\begin{figure*}[t]
	\includegraphics[width=1\textwidth, height=2.5in,trim={1.5cm 9.3cm 14cm 2cm},clip]{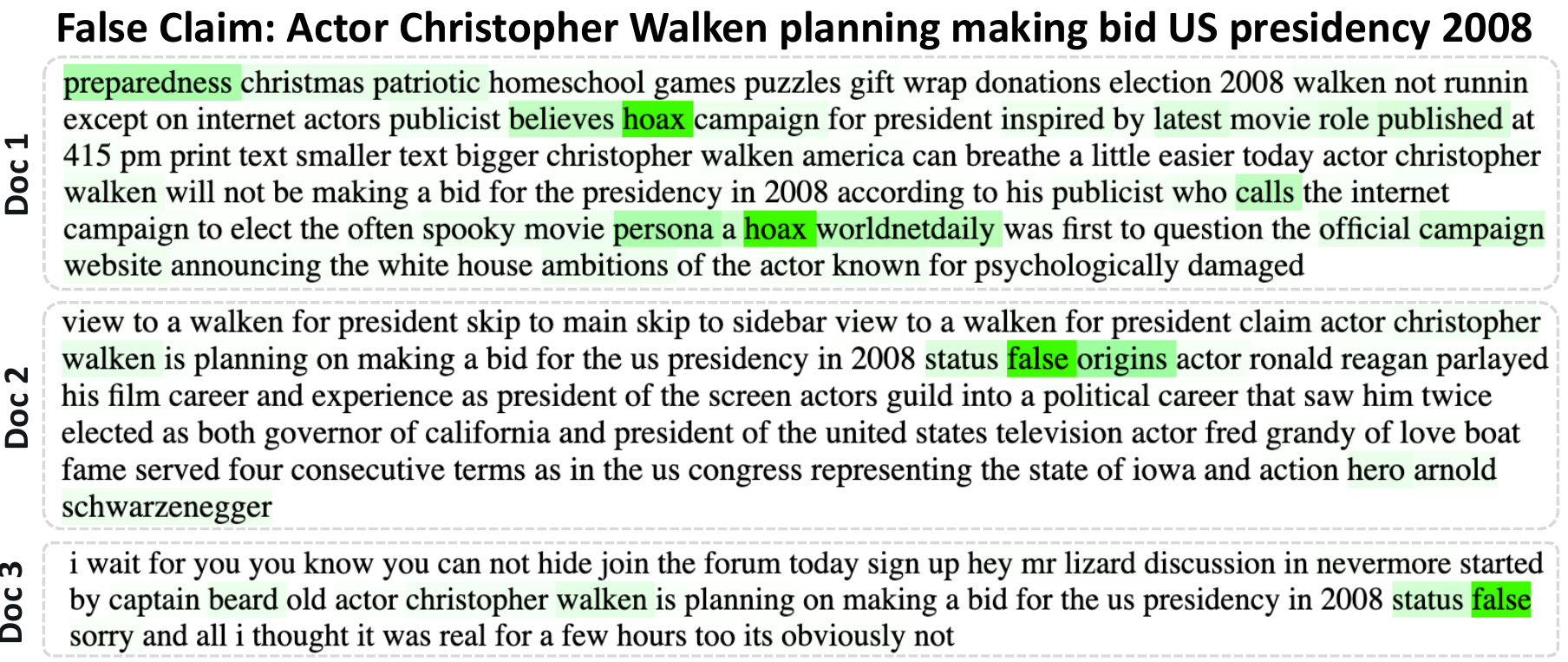}
	\caption{Visualization of attention weights of \textit{the first attention head} on three documents relevant to a false claim in word-level attention layer}
	\label{fig:head1_case_studies}
\end{figure*}
\begin{figure*}[t]
	\includegraphics[width=1\textwidth, height=2.2in,trim={1.5cm 9.3cm 14cm 2cm},clip]{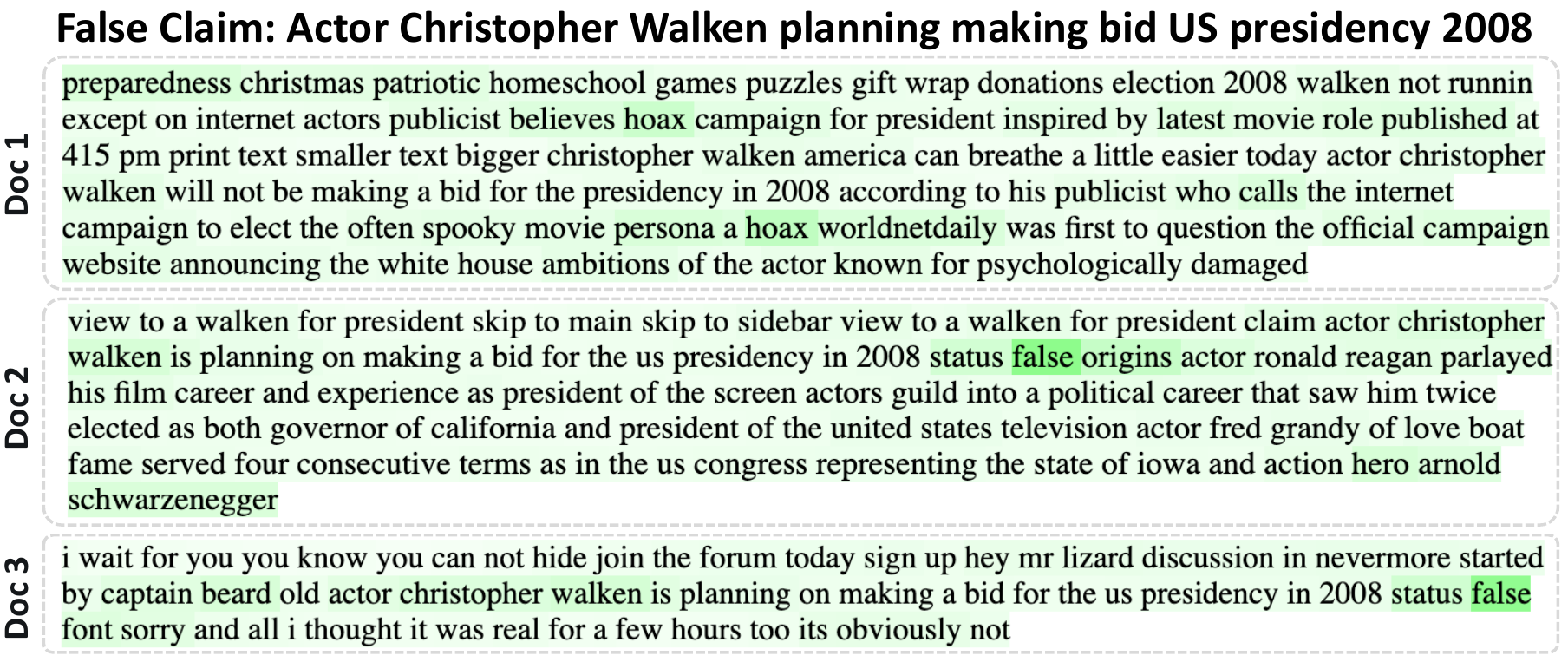}
	\caption{Visualization of attention weights of \textit{the second attention head} on three documents relevant to a false claim in word-level attention layer}
	\label{fig:head2_case_studies}
\end{figure*}
\subsection{Ablation Studies}
\noindent\textbf{Impact of Word Attention and Evidence Attention.}
We study the impact of attention layers on performance of MAC by (1) using only word attention and replacing evidence attention with an average pooling layer on top of documents' representations and (2) using only evidence attention and replacing word attention with an average pooling layer on top of words' representations. As we can see in Table \ref{tbl:ablation_studies_on_word_evd_att}, using only word attention performs much better than using only evidence attention. This is because without downgrading less informative words in evidence, irrelevant information can be captured, leading to low quality representations of evidence. This experiment aligns with our observation that HAN model, which used only evidence attention, did not perform well. When combining both attention mechanisms hierarchically, we consistently achieve best results on two datasets in Table \ref{tbl:ablation_studies_on_word_evd_att}. In particular, the model \textit{Word \& Doc Att} outperformed both \textit{Only Evidence Att} and \textit{Only Evidence Att} significantly with p-value $<$ 0.05. This result indicates that it is crucial to combine word-level attention and document-level attention to improve the performance of evidence-aware fake news detection task.

\noindent\textbf{Impact of Speakers and Publishers on MAC.} To study how speakers and publishers impact performance of MAC, we experiment four models: (1) using text only (Text Only), (2) using text and publishers (Text + Publishers), (3) using text and speakers (Text + Speakers) and (4) using text, publishers and speakers (Text + Pubs + Spkrs). In Table \ref{tbl:ablation_studies}, Text + Publishers has better performance then using only text in both datasets. In PolitiFact, Text + Speakers achieves 2$\sim$3\% improvements over Text + Publishers, indicating that speakers who made claims are crucial to determine verdict of the claims. Finally, using all information (Text + Pubs + Spkrs) helps us achieve the best result in PolitiFact. In Snopes, we omit results of Text + Speakers and Text + Pubs + Spkrs because the dataset does not contain speakers' information. In particular, model \textit{Text + Pubs + Spkrs} outperformed methods \textit{Text Only} and \textit{Text + Publishers} significantly (p-value$<0.05$). Based on these results, we conclude that integrating information of speakers and publishers is useful for detecting misinformation.


\subsection{Impact of the Number of Attention Heads}
In this section, we examine sensitivity of MAC with respect to the number of heads $h_1$ in word attention layer and the number of heads $h_2$ in document attention layer. We vary $h_1$ and $h_2$ in $\{1,2,3,4,5\}$.
Since AUC is less sensitive to any threshold, we report AUC of MAC on two datasets in Fig.~\ref{fig:snopes_auc_h1_h2} and \ref{fig:politifact_auc_h1_h2}. A common pattern we can observe in the two figures is that performance of MAC tends to be better when we increase the number heads $h_1$ in word attention layer while performance of MAC tends to decrease when increasing $h_2$. This phenomenon indicates that word attention is more important than evidence attention. In Snopes, MAC has the best AUC when $h_1 = 5, h_2 = 2$. In PolitiFact, MAC reaches the peak when $h_1=3, h_2=1$.

\subsection{Case Study}

\begin{figure}[t]
	\includegraphics[width=\linewidth, height=1.5in,trim={0.0cm 0.1cm 2.2cm 0.8cm},clip]{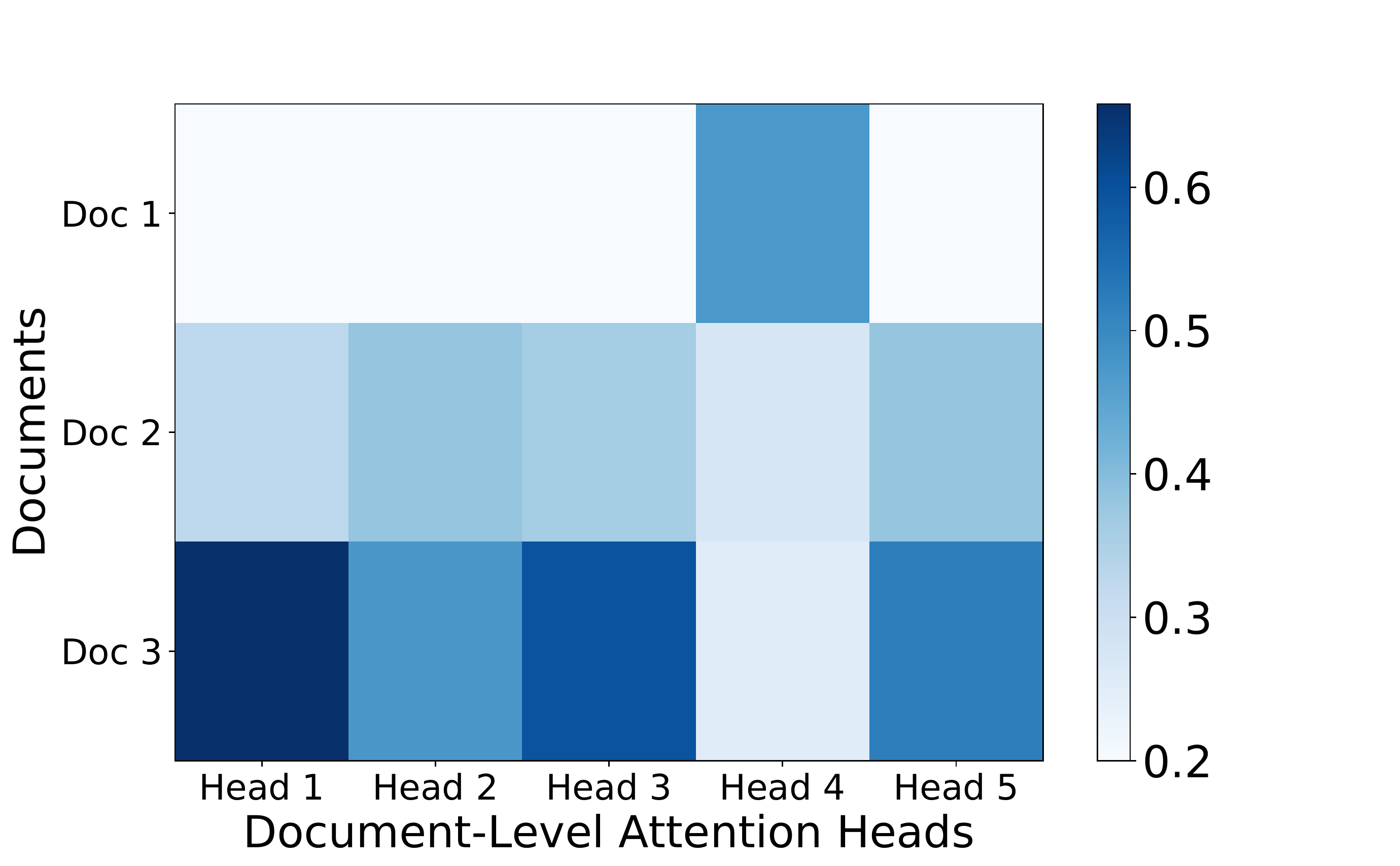}
	\centering
	\caption{Visualization of five attention heads in document-level attention layer for three documents}
	\label{fig:five_document_head}
\end{figure}
To understand how multi-head attention mechanism works, from the testing set, we visualize attention weights on three documents of a false claim \textit{Actor Christopher Walken planning making bid US presidency 2008}. Note, our MAC correctly classifies the claim as fake news. In Fig.~\ref{fig:head1_case_studies} and Fig.~\ref{fig:head2_case_studies}, we show the claim and visualization of two different heads in word attention layer. Note that \citet{popat2018declare}, who released the datasets, already lowercased and removed punctuations. To conduct fair comparison, we directly used the datasets without any additional preprocessing. In Fig.~\ref{fig:head1_case_studies}, attention weights are sparse, indicating that the first attention head focuses on the most important words which determine credibility of the claim (e.g.  \textit{hoax}, \textit{false}). Differently, in Fig.~\ref{fig:head2_case_studies}, the second attention head has more diffused attention weights to capture more useful phrases from documents (e.g. \textit{walken not running}, \textit{its obviously not}).  Moving on to attention heads in evidence attention layer in Fig.~\ref{fig:five_document_head}, we show a heat map where the x-axis is the five heads extracted from evidence attention layer and the y-axis is three documents relevant to the same claim in Fig.~\ref{fig:head1_case_studies} and \ref{fig:head2_case_studies}. As we can see in Fig.~\ref{fig:five_document_head}, \textit{Head 1}, \textit{Head 3} and \textit{Head 5} emphasize on \textit{Doc 3} which contains refuting phrases (e.g. \textit{its obviously not}), while \textit{Head 4} focuses on \textit{Doc 1} which has negating information such as \textit{walken not running}. Both \textit{Doc 1} and \textit{Doc 3} have crucial signals to fact-check the claim. From these analyses, we conclude that heads in word attention layer capture different semantic contributions of words and different heads in document attention layer captures important documents.




\section{Conclusions}
In this paper, we propose a novel evidence-aware model to fact-check textual claims. Our MAC is designed by hierarchically stacking two attention layers. The first one is a word attention layer and the second one is a document attention layer. In both layers, we propose multi-head attention mechanisms to capture different semantic contributions of words and documents.
Our MAC outperforms the baselines significantly with an average increase of 6\% to 9\% over the best results from baselines with a maximum improvements of 18\%. We conduct ablation studies to understand the performance of MAC and provide a case study to show the effectiveness of the attention mechanisms. In future work, we will further examine other data types such as images to improve the performance of our model.

\section*{Acknowledgment}
This work was supported in part by NSF grant CNS-1755536, AWS Cloud Credits for Research, and Google Cloud. Any opinions, findings and conclusions or recommendations expressed in this material are the author(s) and do not necessarily reflect those of the sponsors.
\bibliography{www}
\bibliographystyle{acl_natbib}
\end{document}